# The Distribution of Dependency Distance and Hierarchical Distance in Contemporary Written Japanese and Its Influencing Factors

Linxuan Wang[a*] and Shuiyuan Yu[b]
[a] Department of Japanese, School of Foreign Languages, Renmin University of China, Beijing, China
[b] Institute of Quantitative Linguistics, Faculty of Linguistic Sciences, Beijing Language and Culture University, Beijing, China

## Abstract
To explore the relationship between dependency distance (DD) and hierarchical distance (HD) in Japanese, we compared the probability distributions of DD and HD with and without sentence length fixed, and analyzed the changes in mean dependency distance (MDD) and mean hierarchical distance (MHD) as sentence length increases, along with their correlation coefficient based on the Balanced Corpus of Contemporary Written Japanese. It was found that the valency of the predicates is the underlying factor behind the trade-off relation between MDD and MHD in Japanese. Native speakers of Japanese regulate the linear complexity and hierarchical complexity through the valency of the predicates, and the relative sizes of MDD and MHD depend on whether the threshold of valency has been reached. Apart from the cognitive load, the valency of the predicates also affects the probability distributions of DD and HD. The effect of the valency of the predicates on the distribution of HD is greater than on that of DD, which leads to differences in their probability distributions and causes the mean of MDD to be lower than that of MHD.

**Keywords**: dependency distance, hierarchical distance, valency, trade-off, contemporary written Japanese

## 1. Introduction
As a branch of structural linguistics (Zhou, 1997), dependency grammar, founded by French linguistics Tesnière, has gained increasing attention in recent decades (Tesnière, 1959/2015). According to dependency grammar, words are the basic units of the sentence, and the connections between words form the overall structure of the sentence. There are no words that exist independently of others in a sentence; every word is connected to other words. Each connection involves a head and a dependent, where the dependent depends on the head, thus creating the dependencies between the basic elements of the sentence. Dependency grammar represents the syntactic structure of natural language in a formalized way, endowing it with measurable features. As a result, it has become a key focus of research in quantitative linguistics (Liu et al., 2009; Ouyang & Jiang, 2017; Chen et al., 2022).

DD and HD, proposed within the dependency grammar framework, are two quantitative indicators that reflect different dimensions of natural language. Hudson (1995, p.16) was the first to explicitly define DD as “the distance between words and their parents, measured in terms of intervening words”. For adjacent dependencies, Hudson (1995) considered their DD to be 0. In the sentence “It is remarkable that both of the Siamese twins survived the operation", the DD between *it* and *is* is 0 (p. 17). In contrast, Ferrer-i-Cancho (2004) defined DD as the Euclidean distance between the head and the dependent, suggesting that the DD of adjacent dependencies should be the absolute difference between their positions in the sentence, which equals 1. However, regardless of the calculation method used, DD is an indicator of the linear dimension.

* Correspondence to: Linxuan Wang, linxuanwang01@163.com

On the other hand, as an indicator of the hierarchical dimension, the idea of HD was reflected in Tesnière (1959/2015)'s *stemmas*, but its calculation method was first explicitly defined in Jing and Liu (2015). A sentence can form a corresponding stemma based on the dependencies of its elements. Tesnière (1959/2015) argued that verbs, nouns, adjectives, and adverbs can serve as the central nodes of stemmas. The central node, also known as the root node, is the only element without a head (Ferrer-i-Cancho, 2004, p.1; Liu, 2008, p.163). In all other words, the dependents directly depend on the root node are located on the first layer, the dependents depend on the first-layer words are located on the second layer, and so on, forming a hierarchical stemma. The HD of words in the first layer is 1, and the HD of words in the *n*th layer is *n* (Jing & Liu 2015).

As shown above, both DD and HD are influenced by sentence length. The concept of the average DD proposed by Ferrer-i-Cancho (2004) helps eliminate the influence of sentence length, later referred to as mean dependency distance (MDD) by Liu (2007), a term still in use today. Its calculation involves dividing the sum of all DDs in a sentence by the number of dependencies. Similarly, mean hierarchical distance (MHD) is calculated by dividing the sum of all HDs in a sentence by the number of dependencies it contains (Jing & Liu, 2015).

Previous studies have shown that (M)DD and (M)HD are closely related to human language processing in terms of cognitive resources. Regarding DD, the Dependency Locality Theory (DLT) proposed by Gibson (2000) suggests that "the cost of integrating two elements (such as a head and a dependent, or a pronominal referent to its antecedent) depends on the distance between the two" (p.95). Therefore, the greater the DD, the more cognitive resources are consumed when processing the dependency, resulting in a higher linear complexity of a sentence (Grodner & Gibson, 2005; Ishihara et al., 2020; Chen et al., 2022).

By investigating the dependency treebank of 20 languages, Liu (2008) tested the following three hypotheses: "(1) The human language parser prefers linear orders that minimize the average dependency distance of the recognized sentence (2) There is a threshold that the average dependency distance of most sentences or texts of human languages does not exceed (3) Grammar and cognition combine to keep dependency distance within the threshold" (p.159), and found that adjacent dependencies (DD = 1) account for the highest proportion among all languages. Moreover, the majority of adjacent dependency is also considered a reason for the reduction in MDD.

As for the other dimension, Liu and Jing (2016), based on the spreading activation of a concept proposed by Hudson (2010) and DLT introduced by Gibson (2000), argued that MHD can serve as an indicator of the hierarchical complexity of a sentence. It was found that, like MDD, MHD also tends to be minimized and does not exceed the capacity limit of short-term memory, which is considered 4 (Cowan, 2001). Given this, do the probability distributions of DD and HD, both constrained by cognitive resources, show any similarities?

Regarding the distribution of DD of natural language, Lu and Liu (2016) investigated 30 languages and found that the stretched exponential distribution and the truncated power law distribution are suitable for modeling the probability distributions of DD for short and long sentences, respectively. In a study on a specific language, Liu (2007) investigated the distribution of DD of Chinese and two random treebanks. The results showed that the right-truncated Zeta distribution fits the probability distribution of DD well, which differs significantly from the results of the random treebanks. Moreover, syntax plays a significant role in reducing MDD. In contrast, Ouyang and Jiang (2017) found that the Zipf-Alekseev distribution fits well with the probability distributions of DD of essay data produced by Chinese ESL learners and native speakers, and the parameters of this distribution can predict the English proficiency of Chinese EFL learners at different grade levels. In addition, the probability distribution of DD of Japanese has been shown to follow Zipf's law (Maruyama & Ogino, 1992; Kin, 1996).

Previous studies have focused less on the HD distribution than on the DD distribution, which is related to the chronological order in which DD and HD were proposed. About the hierarchical structure in English, Liu and Jing (2016) found that the distribution of hierarchical numbers (i.e., the maximum HD of the dependency tree plus 1) is positively skewed, indicating that English sentences tend to adopt a flatter hierarchical structure, with the skewness becoming more pronounced as sentence length increases. To investigate whether the position of a word in hierarchical structure affects its ability to govern the words in the next layer, Liu (2017) analyzed the probability distributions of HD in Chinese, Czech, and English corpora. And the study concluded that as the hierarchical level becomes deeper, the number of lower-level words that each upper-level word can govern gradually decreases.

Although DD and HD are indicators of linear and hierarchical dimension respectively, they are not entirely unrelated. Niu et al. (2021) compared the probability distributions of DD of the normal constructions (NC) and the infrequent and hard-to-process syntactic constructions (IHPSC) in English and found no significant differences. However, in terms of the HD distribution, IHPSC is more inclined to avoid an increase in HD compared to NC. Additionally, the larger MDD and smaller MHD of IHPSC suggest that it generally exhibits higher complexity at the linear level, while the tendency to avoid long HDs contributes to a decrease in MHD. This result illustrates the trade-off between MDD, reflecting linear complexity, and MHD, reflecting hierarchical complexity. This trade-off is also found in Czech, as Czech tends to reduce linear complexity, while English tends to avoid an increase in hierarchical complexity (Jing & Liu, 2015).

Previous studies indicate that examining both (M)DD and (M)HD distributions helps us better understand the relationship between linear and hierarchical dimensions of natural language, thereby clarifying how native speakers balance linear complexity and hierarchical complexity to minimize cognitive load in practice. However, the relationship between DD and HD in Japanese has yet to be figured out. Is there a similar trade-off between the two, as seen in English and Czech? If so, what is the underlying cause? To address the above issue, this paper will focus on the following three questions:

(1) What are the differences in the probability distributions of DD and HD in contemporary written Japanese?

(2) How do MDD and MHD change as sentence length (hereafter "SL") increases in contemporary written Japanese?

(3) What are the reasons for the differences in the (M)DD and (M)HD distributions in contemporary written Japanese?

The structure of the manuscript is as follows: Section 2 outlines the materials and methods used in this study and Section 3 presents the findings and discusses possible explanations. The final section summarizes the study and draws conclusions.

## 2. Materials and Methods

### 2.1 Data

The data used in this study are all from the Balanced Corpus of Contemporary Written Japanese (hereafter "BCCWJ")[1] developed by the National Institute for Japanese Language and Linguistics, which contains texts from 13 different registers, with 104 million words in total. To better reflect the usage of contemporary written Japanese, the corpus was constructed by randomly sampling texts from various registers in proportion.

Among the 13 registers in BCCWJ, the internet bulletin board and blog are online resources where anyone can freely post without moderation. Considering that these registers may introduce noise, this study selected 11 other registers for analysis, excluding these two.

In the paid DVD version of BCCWJ (M-XML), the period, as the sentence-ending punctuation, accounts for the highest proportion, comprising 83.30% of all sentences. Japanese closing quotation marks (8.83%), question marks (3.73%), exclamation marks (2.71%), and full-width periods (1.43%) follow in descending order (Yamazaki, 2014). This study selected the periods, question marks, and exclamation marks as the search terms. The exclusion of the closing quotation marks and full-width periods from the search terms is because Japanese closing quotation marks are used not only at the end of sentences but also within sentences to indicate quotation, emphasis on a word, or to highlight a specific part of a sentence. The full-width periods also serve as both a sentence-ending punctuation and a decimal point in the search results. Including the closing quotation marks and full-width periods as search terms and using them for sentence segmentation would introduce many incomplete sentences, thereby affecting syntactic parsing.

To ensure that the retrieval results can cover as much data as possible, we selected the largest window for the search, with a context size of 500 morphemes before and after, and saved the results in a CSV file.[2]

Then we randomly sampled 1% of the final downloaded records for each register. To keep the sentences complete, we concatenated the content following the first sentence-ending punctuation in the preceding context and the content preceding the last sentence-ending punctuation in the following context with the search terms in sequence, followed by data cleaning. We ultimately obtained 320,589 unique sentences.

Finally, the syntactic parsing tools used in this study are *CaboCha* 0.69 (Kudo & Matsumoto, 2002) and the *IPADic* dictionary (Asahara & Matsumoto, 2003), with parsing results represented in terms of *bunsetsu* segments (hereafter "segments"). In the following analysis, we will adopt the term "node" (node) from Tesnière (1959/2015) to abstractly represent Japanese segments. The number of sentences with a specific length, measured in segments, is shown in Figure 1. To cover the majority of sentences and avoid the instability of results due to the limited number of sentences with specific lengths, this study focused on sentences with segments ranging from 2 to 20.

---

[1] https://clrd.ninjal.ac.jp/bccwj/en/index.html (Accessed April 21, 2025).

[2] The version of Chunagon used during the retrieval was 2.7.2, and the data version was 2021.03. The author has obtained full access to the entire dataset by signing a contract with the National Institute for Japanese Language and Linguistics and completing the purchase of the paid version of BCCWJ. The data contained in the paid edition of BCCWJ are identical to those used in Chunagon.

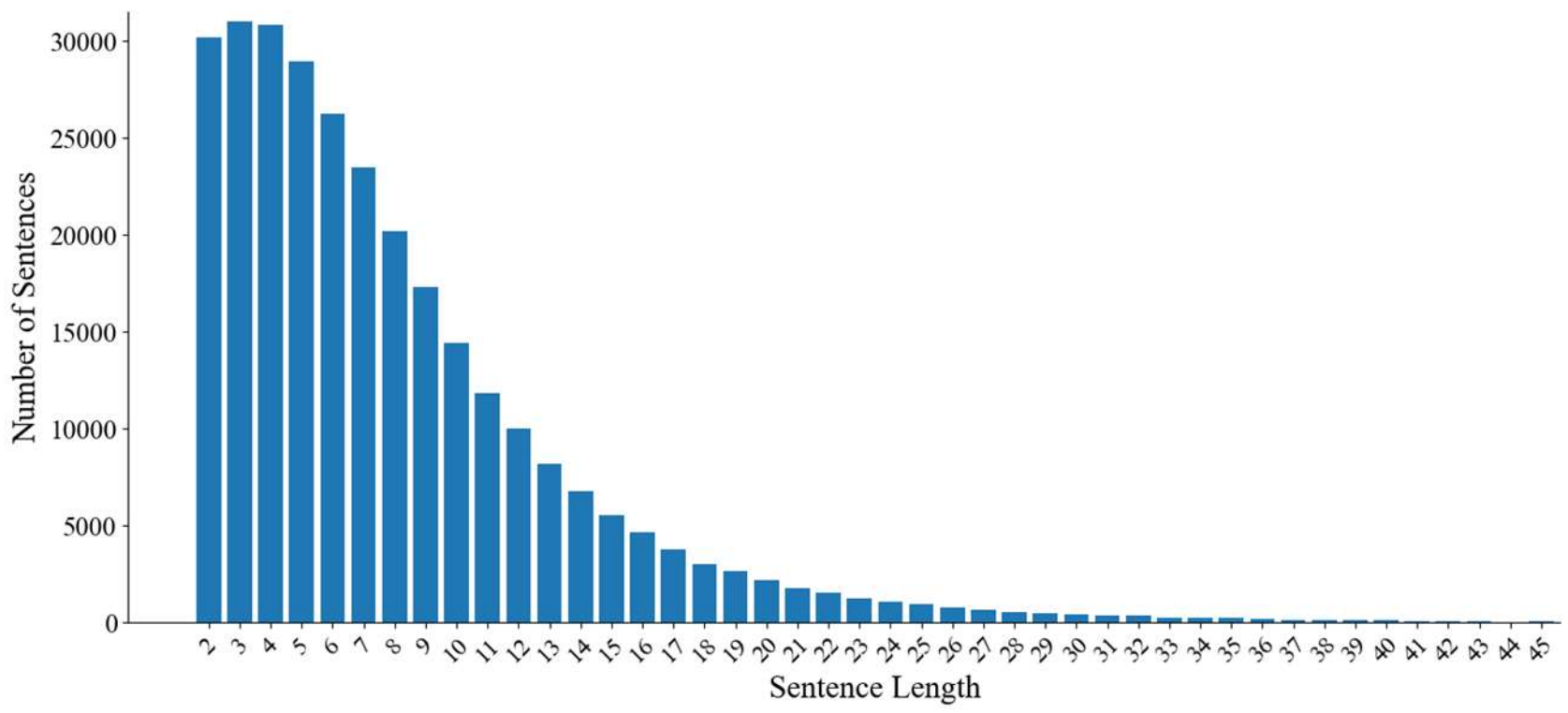

Figure 1. The number of sentences with a specific length (by the segment).

**2.2 The Calculation for Linear and Hierarchical-Level Metrics**

Regarding DD, two calculation methods were mentioned earlier: the number of words between the head and the dependent within the same dependency (Hudson, 1995), or the absolute difference between their positions in the sentence (Ferrer-i-Cancho, 2004). For convenience, this study will adopt the latter method. Taking “この先生は化学の授業を担当する厳しい先生だ(Kono senseiwa kagakuno jyugyouwo tantousuru kibishii senseida)” as an example (the units separated by spaces in parentheses represent segments), the dependency structure of this sentence is represented in Figure 2, where the arrow points from the head of a specific dependency to its dependent.

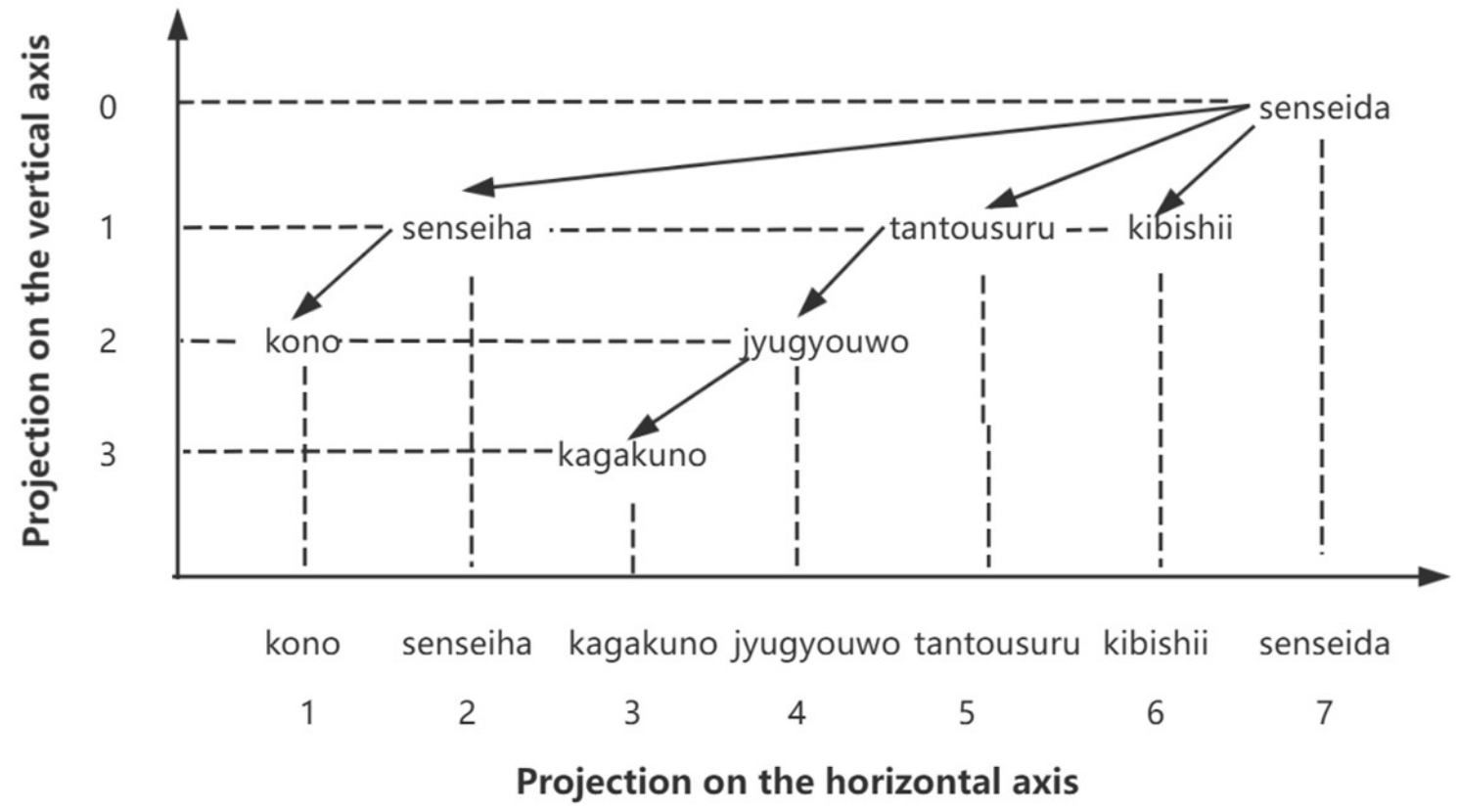

Figure 2. The projection of dependency structure of “この先生は化学の授業を担当する厳しい先生だ（Kono senseiwa kagakuno jyugyouwo tantousuru kibishii senseida）”.

In Figure 2, the number on the vertical axis depends on whether the segment directly depends on the root node, while the number on the horizontal axis represents the position of the segment in the original sentence. The DD of node “kagakuno” is the linear distance to its head “jyugyouwo”, i.e., $|4 - 3| = 1$. The HD is the vertical distance to the root node “senseida”, i.e., 3.

According to Jing and Liu (2015), the MDD of a sentence with $n$ nodes is calculated using formula 1. *DDi* represents the DD of the *i*-th dependency in a sentence. And the MHD of a sentence can be calculated using formula 2, in which *HDi* represents the HD of the *i*-th dependency. In both formulas, $n$-1 represents the number of dependencies in the sentence. Since the root node is not governed by any nodes, the MDD of the example sentence above is $(1 + 5 + 1 + 1 + 2 + 1) / (7 - 1) = 1.8333$, and the MHD is $(2 + 1 + 3 + 2 + 1 + 1) / (7 - 1) = 1.667$.

$$\text{MDD (the sentence)} = \frac{1}{n-1}\sum_{i=1}^{n-1}|DD_i| \quad (1)$$

$$\text{MHD (the sentence)} = \frac{1}{n-1}\sum_{i=1}^{n-1}HD_i \quad (2)$$

## 3. Results and Discussion

### 3.1 The Probability Distributions of DD and HD

The probability distributions of DD and HD without SL fixed are shown in Figure 3. The most frequent values for both DD and HD are DD = 1 and HD = 1, indicating that both MDD and MHD tend to be minimized (Liu et al., 2017). However, the proportion of dependencies with HD = 1 (33.31%) is approximately half of the proportion of dependencies with DD = 1 (64.54%).

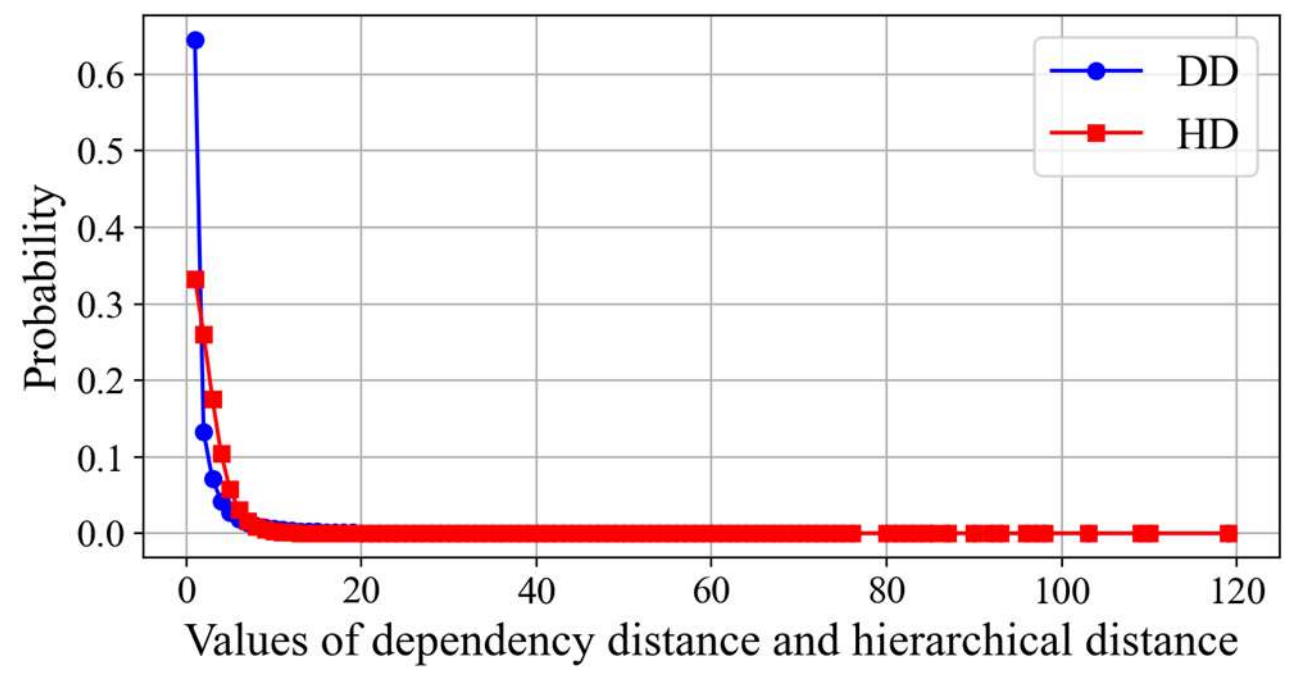

Figure 3. The probability distributions of DD and HD when SL = 2 to 20.

As an increase in SL leads to higher DD and HD (Ferrer-i-Cancho, 2004; Jiang & Liu, 2015; Niu et al., 2021), which in turn raise MDD and MHD, we analyzed the probability distributions of DD and HD with SL fixed in contemporary written Japanese (Figure 4(a) and (b)), along with their entropy (Figure 5). To present the trend of the data clearly and avoid the randomness caused by small sample sizes, we output the probability distributions of both from SL = 5 to 30 in intervals of 5. It can be observed that although the uncertainty of the probability distributions of DD and HD increases with SL, the probability distributions of DD values remain highly consistent across different SLs. This indicates that the increase in SL allows for more possible values of DD, but the proportion of each DD value does not significantly change with the increase in SL. The proportion of the most frequent dependency (DD = 1) remains around 63%. As SL increases, the proportion of HD = 1 gradually decreases, while the HD value with the highest probability increases.

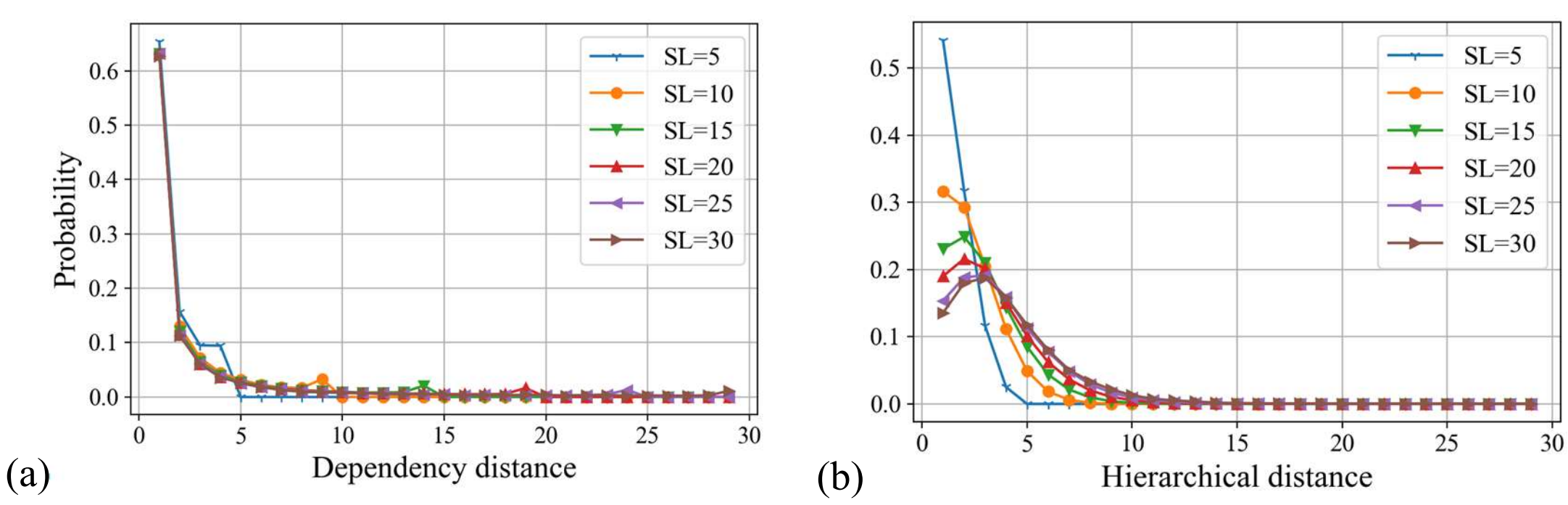

Figure 4. The probability distributions of (a) DD and (b) HD with SL fixed.

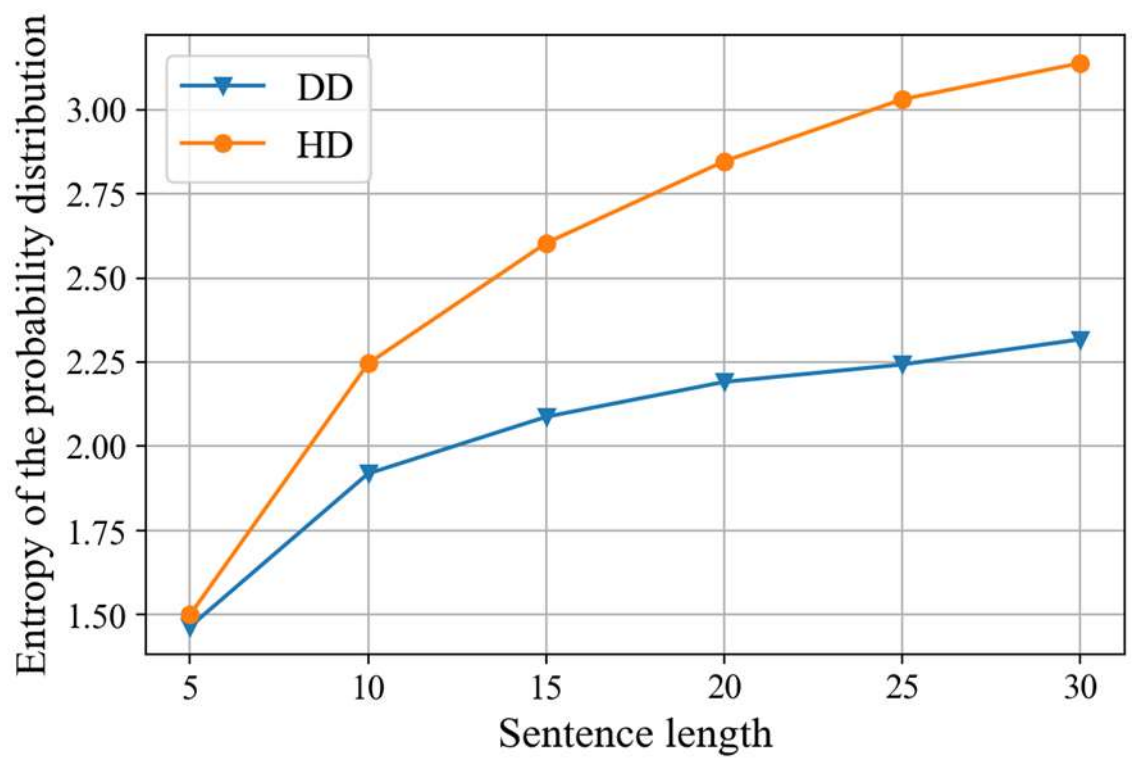

Figure 5. The entropy of the probability distributions of DD and HD with SL fixed.

### 3.2 The Connection Between MDD, MHD and SL

Then we calculated the means of MDD and MHD in contemporary written Japanese when SL ranges from 2 to 20 (Figure 6(a)). As SL increases, the means of both MDD and MHD gradually increase, which is consistent with the findings of previous studies (Ferrer-i-Cancho, 2004; Liu & Jing, 2016). MDD and MHD intersect between SL = 5 and 6. Before the intersection, MDD is greater than MHD, and after the intersection, their relationship reverses.

Imada (2023, p.14) suggested that the appearance of the hierarchical structure in Japanese sentences helps to reduce MDD. It can be inferred that, for the same number of segments (i.e., SL), the changes in MDD and MHD are interrelated. Therefore, we calculated the Spearman's correlation coefficients of MDD and MHD with SL fixed (Figure 6(b)).

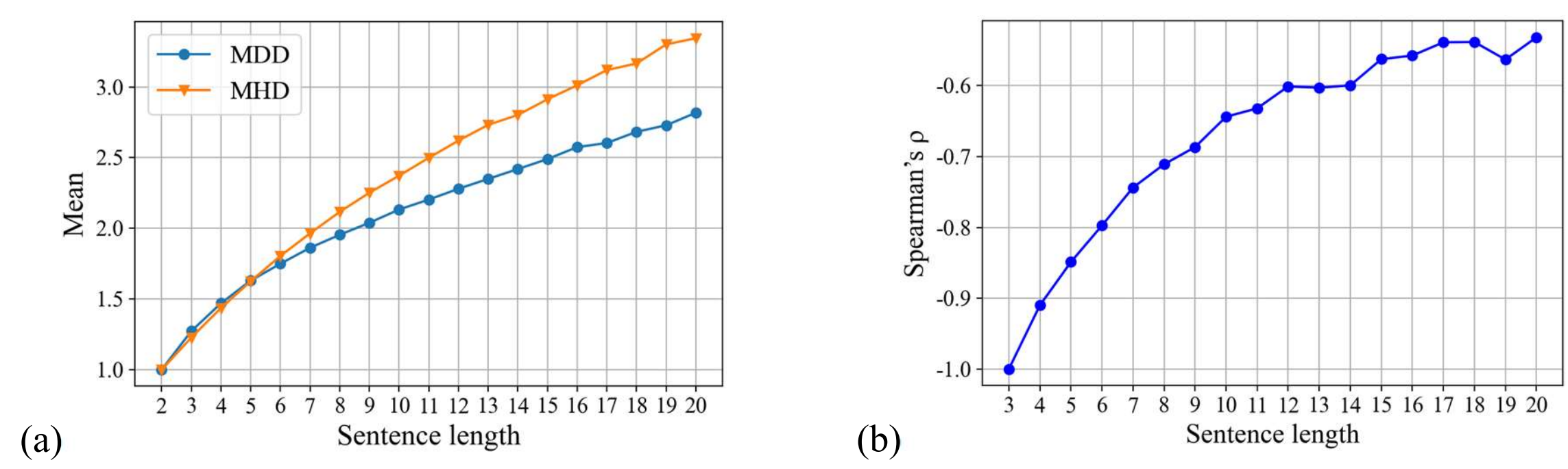

Figure 6. (a) The means of MDD and MHD increase with SL. (b) The Spearman's correlation coefficients of MDD and MHD is significantly negative when SL = 3 to 20.

Since both MDD and MHD are always equal to 1 when SL is 2, with no variance, it is not appropriate to calculate the correlation coefficients in this case. In Figure 6(b), MDD is significantly negatively correlated with MHD when SL = 3 to 20. And as SL increases, the negative correlation between MDD and MHD gradually declines.

### 3.3 The Relationship Between Valency, DD and HD

In Section 3.1, an increase in the entropy of the probability distributions of DD and HD is observed as SL increases. However, the proportion of nodes with DD = 1 to DD = 29 do not fluctuate significantly with the increase in SL. The proportion of DD = 1 remains around 63%. It can be inferred that due to the limitations of cognitive resources, native speakers of Japanese tend to use more adjacent dependencies to minimize MDD regardless of SL. Although both MDD and MHD tend to be minimized without SL fixed, the proportion of HD = 1 is half of that for DD = 1. Additionally, the proportion of HD = 1 decreases as SL increases, with the reduced proportion shifting to other HD values, causing the HD value with the highest proportion to gradually increase. These facts suggest that the factors influencing the probability distributions of HD and DD are not entirely the same.

Based on the result in Section 3.2, where the means of MDD and MHD intersect between SL = 5 and 6, we argue that, in addition to the cognitive load, the valency of the predicates also influences the distribution of HD. Valency refers to the ability of a word to combine with other words in a broad sense (Liu & Feng, 2007). And it is considered that verbs, adjectives, and nouns in contemporary written Japanese can serve as the predicate (Muraki, 2007, p.11; Sanada, 2016, p.261; Ohkawa, 2023, p.158), with verbs having a maximum of four segments, and adjectives a maximum of two segments (The Japanese Descriptive Grammar Research Group, 2009).

If the maximum valency of the predicate is assumed, then in a sentence where verb serves as the predicate, up to four nodes can directly depend on the root node, in which case SL = 5. Therefore, before the predicate reaches its threshold of valency (i.e., when the nodes are less than or equal to 5), all nodes except the root node in the sentence can directly depend on the root node, with their HD equal to 1. However, since humans must output natural language in a linear sequence, nodes that directly depend on the root node will lead to relatively longer DDs, which is why the mean of MDD is greater than that of MHD before the intersection.

After the predicate reaches the maximum of valency (i.e., when the number of nodes is greater than or equal to 6), the root node can no longer govern all other nodes. As a result, the nodes that cannot be governed by the predicate will shift to a deeper hierarchy with HD ≥ 2, leading to a continued increase in the mean of MHD compared to that of MDD. Meanwhile, the valency of the predicate also provides a good explanation for the differences between the distributions of HD and DD discussed in Section 3.1.

On the other hand, the significant negative correlation between MDD and MHD with SL fixed also reflects the trade-off relation between them. Because the valency of the predicate limits the number of segments that can directly depend on the root node, it slows down the growth of MDD while promoting the growth of MHD at the same time. It can be assumed that there is an interaction between linear and hierarchical complexity in contemporary written Japanese, and this interaction largely stems from the valency of the predicate. Therefore, we may conclude that in addition to the cognitive load mentioned in previous studies, the influence of the valency on the (M)DD and (M)HD distributions is also non-negligible.

To further explore the impact of the valency of the predicates on the distributions of DD and HD, we focused on the monovalent verbs to quadrivalent verbs listed in The Japanese Descriptive Grammar Research Group (2009) and analyzed how the average number of nodes with DD = 1 and HD = 1 changes with SL when the valency of the predicates is fixed (Figure 7(a) and (b)).

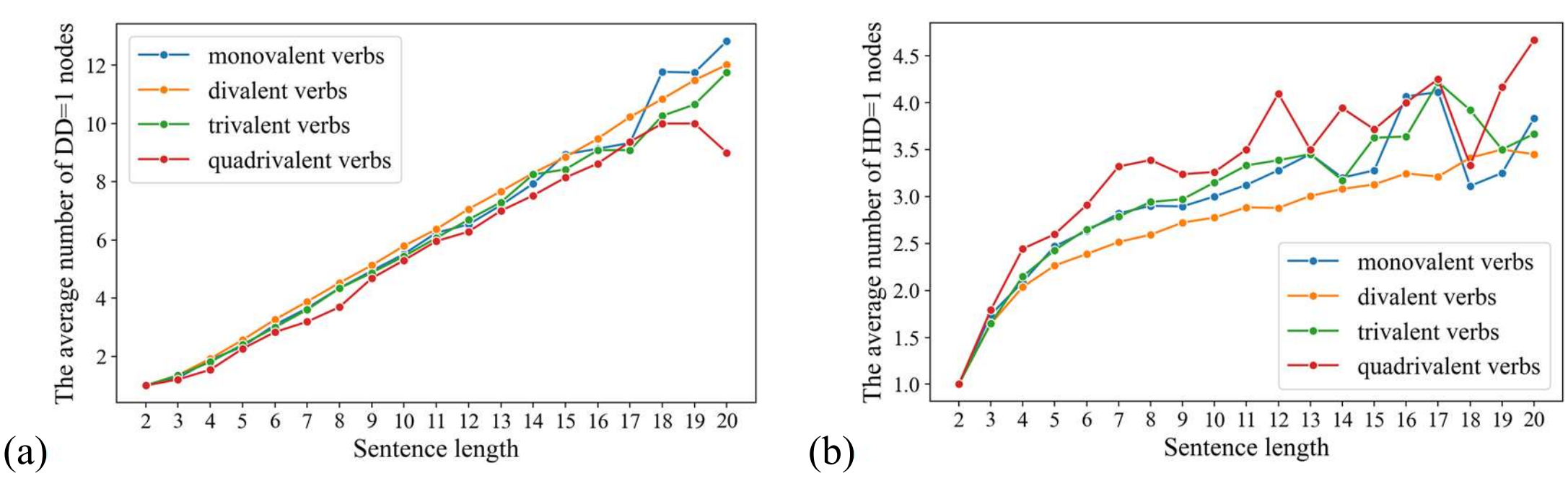

Figure 7. The average number of nodes with (a) DD = 1 and (b) HD = 1 for different SLs.

When the valency is held constant, the average number of nodes with DD = 1 increases linearly with SL (Figure 7(a), with all $R^2$ greater than 0.97 in Table 1). In general, as valency increases, the average number of nodes with DD = 1 decreases, while the average number of nodes with HD = 1 increases. In Figure 7(b), the data show a downward-curving trend. When the valency of the predicate ranges from 1 to 4, the $R^2$ of the linear regression model between the logarithm of SL and the number of nodes with HD = 1 is greater than 0.86 in all cases (Table 2). As SL increases, the number of nodes with HD = 1 gradually approaches 4. It is assumed that the average number of nodes with HD = 1 for certain SL exceeds 4 due to the different definitions of valency. In the Japanese Descriptive Grammar Research Group (2009), only the segments containing nominal component(s) and case(s) are considered as the valency of verbs. However, adverbial segments in contemporary Japanese can also directly modify verbs (Noda, 2006, p.191), and CaboCha also analyzes the head of such adverbial segments as the verb functioning as the predicate (with only one verb in a sentence). Therefore, if we define valency as the number of nodes that directly depend on the root node in the CaboCha parsing results, the average number of nodes with HD = 1 would be ≥ 4.

Table 1. The results of linear regression models between SL and the number of nodes with DD = 1 when the valency is held constant.

| Valency of verbs | N | *x* | intercept | model | adjusted $R^2$ |
|---|---|---|---|---|---|
| 1 | 19 | $0.6479^{***}$<br>(0.018) | $-0.8269^{***}$<br>(0.221) | y = 0.6479x−0.8269 | 0.986 |
| 2 | 19 | $0.6267^{***}$<br>(0.003) | $-0.4861^{***}$<br>(0.039) | y = 0.6267x−0.4861 | 1.000 |
| 3 | 19 | $0.5929^{***}$<br>(0.009) | $-0.4487^{***}$<br>(0.106) | y = 0.5929x−0.4487 | 0.996 |
| 4 | 19 | $0.5440^{***}$<br>(0.020) | −0.3183<br>(0.240) | y = 0.5440x−0.3183 | 0.977 |

Table 2. The results of linear regression models between the logarithm of SL and the number of nodes with HD = 1 when the valency is held constant.

| Valency of verbs | N | log($x$) | intercept | model | adjusted $R^2$ |
|---|---|---|---|---|---|
| 1 | 19 | 1.0753***<br>(0.101) | 0.5643***<br>(0.234) | y = 1.0753log(x)+0.5643 | 0.862 |
| 2 | 19 | 0.9714***<br>(0.032) | 0.5578***<br>(0.075) | y = 0.9714log(x)+0.5578 | 0.980 |
| 3 | 19 | 1.1671***<br>(0.072) | 0.4328***<br>(0.167) | y = 1.1671log(x)+0.4328 | 0.935 |
| 4 | 19 | 1.2713***<br>(0.110) | 0.4890**<br>(0.254) | y = 1.2713log(x)+0.4890 | 0.881 |

*Note: In Table 1 and Table 2, N represents the sample size, $x$ represents SL, *** indicates a $p < 0.05$, ** indicates a $p \geq 0.05$ but $< 0.1$, and the values in parentheses represent the standard error.

However, our definition of valency only affects the threshold to which the average number of nodes with HD = 1 converges; it does not change the fact that such a threshold exists—unlike the case of DD = 1, where the average number of nodes tends to increase in a linear pattern. In addition, considering that the proportion of dependencies with DD = 1 is about twice that of those with HD = 1 without SL fixed (Figure 3), and that the proportion of HD = 1 gradually decreases with increasing SL while the proportion of DD = 1 remains relatively stable with SL fixed (Figure 4(a) and (b)), it can be inferred that while both the distributions of DD and HD are constrained by cognitive resources, the distribution of HD is more strongly influenced by the valency of the predicate. As a result, the mean of MDD tends to be minimized more than that of MHD (Figure 6(a)). The valency of the predicate is also a contributing factor to the trade-off observed between MDD and MHD in contemporary written Japanese. Therefore, it can be argued that native speakers regulate the linear complexity and hierarchical complexity of Japanese through the valency of the predicates. However, whether the cognitive load when both dimensions reach a relative balance, regulated by the valency of the predicates, is smaller than the cognitive load when only prioritizing the increase in one dimension's complexity, that is, whether the cognitive load at balance is the optimal solution for native speakers, still requires verification through experiments in psychology and cognitive neuroscience.

## 4. Conclusion

To explore the relationship between DD and HD in contemporary written Japanese, this study analyzed data from BCCWJ and addressed the following three research questions: (1) What are the differences in the probability distributions of DD and HD in Japanese? (2) How do MDD and MHD change as SL increases in Japanese? (3) What are the reasons for the differences in the (M)DD and (M)HD distributions? It was found that (1) the proportion of DD = 1 is twice that of HD = 1 without SL fixed. When SL is fixed, there is no significant change in the proportion of each DD as SL increases. In contrast, the proportion of HD = 1 decreases with increasing SL, while the HD with the highest proportion increases; (2) As SL increases, both MDD and MHD gradually increase, intersecting between SL = 5 and 6; (3) The main factor contributing to the differences in the probability distributions of DD and HD of Japanese is the valency of the predicates. Both probability distributions of DD and HD are constrained by cognitive resources, but the distribution of HD is more strongly influenced by the valency of the predicates than that of DD. This inference is reflected in the following results: (1) Although both MDD and MHD show a general tendency toward minimization in their probability distributions, the proportion of nodes with HD = 1 cannot reach the same level as those with DD = 1 (when SL is not fixed); (2) As SL increases, the proportion of nodes with HD = 1 decreases; (3) When the valency is held constant, the number of nodes with HD = 1 shows a downward curve and gradually approaches the upper limit of the valency of the predicate—that is, the maximum number of segments that can directly depend on the predicate; (4) The mean of MDD tends to be lower than that of MHD.

Furthermore, this study suggested that the valency of the predicates is the underlying reason for the trade-off between MDD and MHD in Japanese, and that native speakers regulate the linear and hierarchical complexity of the language through the valency of the predicates. The valency of the predicates limits the number of nodes that can directly depend on the root node. Therefore, before the threshold of valency has been reached, i.e., the root node is still capable of forming direct dependencies with additional nodes, increasing DD (>1) and consistently low HD (=1) will result in MDD being greater than MHD. However, once the maximum valency of the predicates is reached (i.e., SL exceeds 5), nodes that exceed the valency limit can no longer directly depend on the root node, leading to increasing HD. As a result, MHD becomes greater than MDD.

Due to space limitations, this study has not addressed the reason for the weakening negative correlation between MDD and MHD as SL increases, which will be explored in future research. It is also worth further investigating whether factors other than the valency of the predicate also contribute to the trade-off between (M)DD and (M)HD.

## Acknowledgements

We are grateful to Haitao Liu from Fudan University for his constructive suggestions, and Linxuan Wang would like to thank her family and friends, including Man Li, Ya Luo, Jing Ouyang and Yifan Tian (listed alphabetically by surname) for their selfless assistance and unconditional support.

## References

Asahara, M., & Matsumoto, Y. (2003). *ipadic version 2.7.0 User's Manual*. Nara Institute of Science and Technology. https://paperzz.com/doc/7739961/ipadic-version-2.7.0-user-s-manual (Accessed April 24, 2025).

Chen, R., Deng, S., & Liu, H. (2022). Syntactic Complexity of Different Text Types: From the Perspective of Dependency Distance Both Linearly and Hierarchically. *Journal of Quantitative Linguistics*, *29*(4), 510–540. https://doi.org/10.1080/09296174.2021.2005960

Cowan, N. (2001). The magical number 4 in short-term memory: A reconsideration of mental storage capacity. *Behavioral and Brain Sciences*, *24*(1), 87–114. https://doi.org/10.1017/S0140525X01003922

Ferrer-i-Cancho, R. (2004). Euclidean distance between syntactically linked words. *Physical Review. E, Statistical, Nonlinear, and Soft Matter Physics*, *70 5 Pt 2*, 056135.

Gibson, E. (2000). The dependency locality theory: A distance-based theory of linguistic complexity. In A. P. Marantz, Y. Miyashita, & W. O'Neil (Eds.), *Image, language, brain* (pp. 95–126). The MIT Press.

Grodner, D., & Gibson, E. (2005). Consequences of the serial nature of linguistic input for sentential

complexity. *Cognitive Science*, *29*(2), 261–290. https://doi.org/10.1207/s15516709cog0000_7
Hudson, R. (1995). Measuring syntactic difficulty. Unpublished paper. Available at http://dickhudson.com/wp-content/uploads/2013/07/Difficulty.pdf (Accessed April 22, 2025).
Hudson, R. (2010). *An introduction to word grammar*. Cambridge University Press.
Imada, M. (2023). Distribution of sentence length and dependency distance in children's compositions: Characteristics of natural language and variations in language development. *F1000Research*, *12*, 379. https://doi.org/10.12688/f1000research.132383.1
Ishihara, H., Iribe, Y., & Kitaoka, N. (2020). Goi to kakari-uke kōzō ni chakumoku shita zatsudan taiwa kara no ninchishō keikō kenshutsu [Detection of dementia tendencies from casual conversations focusing on lexical and dependency structures]. *Dai 82-kai zenkoku taikai kōen ronbunshū* [Proceedings of the 82nd National Convention of IPSJ], *2020*(1), 459–460.
Jiang, J., & Liu, H. (2015). The effects of sentence length on dependency distance, dependency direction and the implications–based on a parallel English–Chinese dependency treebank. *Language Sciences*, *50*, 93–104. https://doi.org/10.1016/j.langsci.2015.04.002
Jing, Y., & Liu, H. (2015). Mean Hierarchical Distance Augmenting Mean Dependency Distance. *Proceedings of the Third International Conference on Dependency Linguistics (Depling 2015)*, 161–170. https://aclanthology.org/W15-2119
Kin, M. (1996). Bunsetsu no kakari-uke kyori no tōkei bunseki [Statistical analysis of dependency distance between Japanese phrases]. *Shakai Jōhō* [Social Information], *5*(2), 1–11.
Kudo, T., & Matsumoto, Y. (2002). Japanese dependency analysis using cascaded chunking. *COLING-02: The 6th Conference on Natural Language Learning 2002 (CoNLL-2002)*. CoNLL 2002. https://aclanthology.org/W02-2016
Liu, H. (2007). Probability distribution of dependency distance. *Glottometrics*, *15*, 1–12.
Liu, H. (2008). Dependency distance as a metric of language comprehension difficulty. *Journal of Cognitive Science*, *9*(2), 159–191. https://doi.org/10.17791/JCS.2008.9.2.159
Liu, H. (2017). Juzi jiegou cengji de fenbu guilü [Distribution patterns of hierarchical sentence structure]. *Foreign Language Teaching and Research*, *49*(3), 345–352+479.
Liu, H., & Feng, Z. (2007). Ziran yuyan chuli de gaïlü peijia moshi lilun [The probabilistic valency model theory in natural language processing]. *Linguistic Sciences*, *3*, 32–41.
Liu, H., & Jing, Y. (2016). Yingyu juzi cengji jiegou jiliang fenxi [Quantitative analysis of hierarchical structures in English sentences]. *Journal of Foreign Languages*, *39*(6): 2–11.
Liu, H., Hudson, R., & Feng, Z. (2009). Using a Chinese treebank to measure dependency distance. *Corpus Linguistics and Linguistic Theory*, *5*(2), 161–175. https://doi.org/10.1515/CLLT.2009.007
Liu, H., Xu, C., & Liang, J. (2017). Dependency distance: A new perspective on syntactic patterns in natural languages. *Physics of Life Reviews*, *21*, 171–193. https://doi.org/10.1016/j.plrev.2017.03.002
Lu, Q., & Liu, H. (2016). Yicun juli fenbu you guilü ma? [Is there a pattern in the distribution of dependency distance?]. *Journal of Zhejiang University (Humanities and Social Sciences Edition)*, *46*(4), 63–76.
Maruyama, H., & Ogino, S. (1992). Nihongo ni okeru bunsetsukan kakari-uke kankei no tōkeiteki seishitsu [Statistical properties of dependency relations between phrases in Japanese]. *Zenkoku taikai kōen ronbunshū, dai 45-kai (Jinkō chinō oyobi ninchi kagaku)* [Proceedings of the 45th

National Convention (Artificial Intelligence and Cognitive Science)], 173–174.
Muraki, S. (2007). Nihongo no setsu no ruikei [Types of clauses in Japanese]. *Dōshisha joshi daigaku gakujutsu kenkyū nenpō* [Annual Journal of Academic Research of Doshisha Women's College of Liberal Arts], *58*, 9–17. https://doi.org/10.15020/00000391
Nihongo Kijutsu Bunpō Kenkyūkai [The Japanese Descriptive Grammar Research Group]. (2009). *Gendai nihongo bunpō 2: Dai 3-bu kaku to kōbun, dai 4-bu voisu* [Modern Japanese grammar, Vol. 2: Part 3 Case and syntax, Part 4 Voice]. Tokyo: Kuroshio Shuppan.
Niu, R., Wang, Y., & Liu, H. (2021). The properties of rare and complex syntactic constructions in English. A corpus-based comparative study. *Proceedings of the Second Workshop on Quantitative Syntax (Quasy, SyntaxFest 2021)*, 74–83.
Noda, H. (2006). Go no junjo, seibun no junjo, bun no junjo: Junjo no jiyūdo to junjo no dōki [Word order, constituent order, and sentence order: Degrees of freedom and motivations for ordering]. T. Masuoka, H. Noda, & T. Moriyama (Eds.), *Nihongo bunpō no shinchihei 1: Keitai, jojutsu naiyō hen* [New Horizons in Japanese Grammar, Vol. 1: Morphology and Descriptive Content] (pp. 179–199). Tokyo: Kuroshio Shuppan.
Ohkawa, K. (2023). *Kodai nihongo buntai no keiryōteki kenkyū* [Quantitative studies on ancient Japanese writing styles]. Tokyo: Musashino Shoin.
Ouyang, J., & Jiang, J. (2017). Can the Probability Distribution of Dependency Distance Measure Language Proficiency of Second Language Learners? *Journal of Quantitative Linguistics*, *28*(4), 295–313. https://doi.org/10.1080/09296174.2017.1373991
Sanada, H. (2016). The Menzerath-Altmann Law and Sentence Structure. *Journal of Quantitative Linguistics*, *23*(3), 256–277. https://doi.org/10.1080/09296174.2016.1169850
Tesnière, L. (2015). *Elements of Structural Syntax* (original work published in 1959, translated by Timothy Osborne and Sylvain Kahane). Amsterdam: John Benjamins.
Yamazaki, M. (2014). Influence of word unit and sentence length on the ratio of parts of speech. *Proceedings of the 5th Workshop on Corpus-Based Japanese Linguistics*, 233–242.
Zhou, L. (1997). Jin wushi nian lai yuyanxue de fazhan (shang) [The development of linguistics over the past fifty years (Part 1)]. *Foreign Language Teaching and Research*, *3*, 24-30+83.